\documentclass[10pt,twocolumn,letterpaper]{article}
\pdfoutput=1
\usepackage{iccv,overpic}
\usepackage{times}
\usepackage{epsfig}
\usepackage{graphicx}
\usepackage{amsmath}
\usepackage{amssymb}
\usepackage[pagebackref=true,breaklinks=true,letterpaper=true,colorlinks,bookmarks=false]{hyperref}
\iccvfinalcopy % *** Uncomment this line for the final submission

\newcommand{\myparagraph}[1]{\vspace{0.1em}\noindent\textbf{#1}}
\usepackage{pifont}
\usepackage{multirow}
\usepackage{color}
\usepackage{mathtools}% http://ctan.org/pkg/mathtools
\usepackage{arydshln}
\newcommand{\cmark}{\ding{51}}
\newcommand{\xmark}{\ding{55}}
\usepackage{float}
\def\iccvPaperID{6670} % *** Enter the ICCV Paper ID here

\ificcvfinal\fi
\begin{document}
%------------------------------------------------------------------------------------
\title{Self-Regulation for Semantic Segmentation}
%------------------------------------------------------------------------------------
\author{
Dong Zhang$^{1}$ \quad Hanwang Zhang$^{2}$ \quad Jinhui Tang$^{1}$\thanks{Corresponding author.} \quad Xian-Sheng Hua$^{3}$ \quad Qianru Sun$^{4}$ \\
\small$^{1}$School of Computer Science and Engineering, Nanjing University of Science and Technology\\
\small$^{2}$Nanyang Technological University \quad
$^{3}$Damo Academy, Alibaba Group \quad
$^{4}$Singapore Management University
}
\maketitle
%------------------------------------------------------------------------------------
%\twocolumn[{%
%\renewcommand\twocolumn[1][]{#1}%
%\maketitle
%\begin{figure}[H]
%\hsize=\textwidth % cvpr需要
%\centering
%\includegraphics[width=.99\textwidth]{latex/tabs_and_figs/fig1.pdf}
%\caption{Our inspections on two failure cases of  the baseline method DeepLab-v2~\cite{chen2017deeplab}, in (a) and (f). The prediction in (a) has the flaw of missing ``horse legs''. The visualization of shallow-layer features in (b) shows that the model indeed attends to ``horse legs'' but also to background noises. Applying our method SR, in (d), reduces noises while maintaining the desirable attention on ``horse legs''. In addition, deep-layer features in (c) and (e) also prove the effectiveness of SR. The prediction in (f) has the confusion between object classes (\ie, ``horse'' and ``cow''). The revision in (g) is obtained by adding class-level cross-entropy losses (\ie, using our MEA loss on DeepLab-v2).}
%\label{fig1}
%\end{figure}
%}]
%------------------------------------------------------------------------------------
\begin{abstract}
In this paper, we seek reasons for the two major failure cases in Semantic Segmentation (SS): 1) missing small objects or minor object parts, and 2) mislabeling minor parts of large objects as wrong classes. We have an interesting finding that Failure-1 is due to the underuse of detailed features and Failure-2 is due to the underuse of visual contexts. To help the model learn a better trade-off, we introduce several Self-Regulation (SR) losses for training SS neural networks. By ``self'', we mean that the losses are from the model \textit{per se} without using any additional data or supervision. By applying the SR losses, the deep layer features are regulated by the shallow ones to preserve more details; meanwhile, shallow layer classification logits are regulated by the deep ones to capture more semantics. We conduct extensive experiments on both weakly and fully supervised SS tasks, and the results show that our approach consistently surpasses the baselines. We also validate that SR losses are easy to implement in various state-of-the-art SS models, \eg, SPGNet~\cite{cheng2019spgnet} and OCRNet~\cite{YuanCW19}, incurring little computational overhead during training and none for testing\footnote{The code is available at: https://github.com/dongzhang89/SR-SS}.
% In this paper, we seek reasons for the two major failure cases in Semantic Segmentation (SS): 1) missing small objects or minor object parts, and 2) mislabeling minor parts of large objects as wrong classes. We have an interesting finding that Failure-1 is due to the underuse of detailed features and Failure-2 is due to the underuse of visual contexts. To help the model learn a better trade-off, we introduce several Self-Regulation (SR) losses for training SS neural networks. By "self", we mean that the losses are from the model per se without using any additional data or supervision. By applying the SR losses, the deep layer features are regulated by the shallow ones to preserve more details; meanwhile, shallow layer classification logits are regulated by the deep ones to capture more semantics. We conduct extensive experiments on both weakly and fully supervised SS tasks, and the results show that our approach consistently surpasses the baselines. We also validate that SR losses are easy to implement in various state-of-the-art SS models, e.g., SPGNet and OCRNet, incurring little computational overhead during training and none for test.
\end{abstract}
%------------------------------------------------------------------------------------
\section{Introduction} \label{sec_intro}
\begin{figure}[t]
\centering
\includegraphics[width=.46\textwidth]{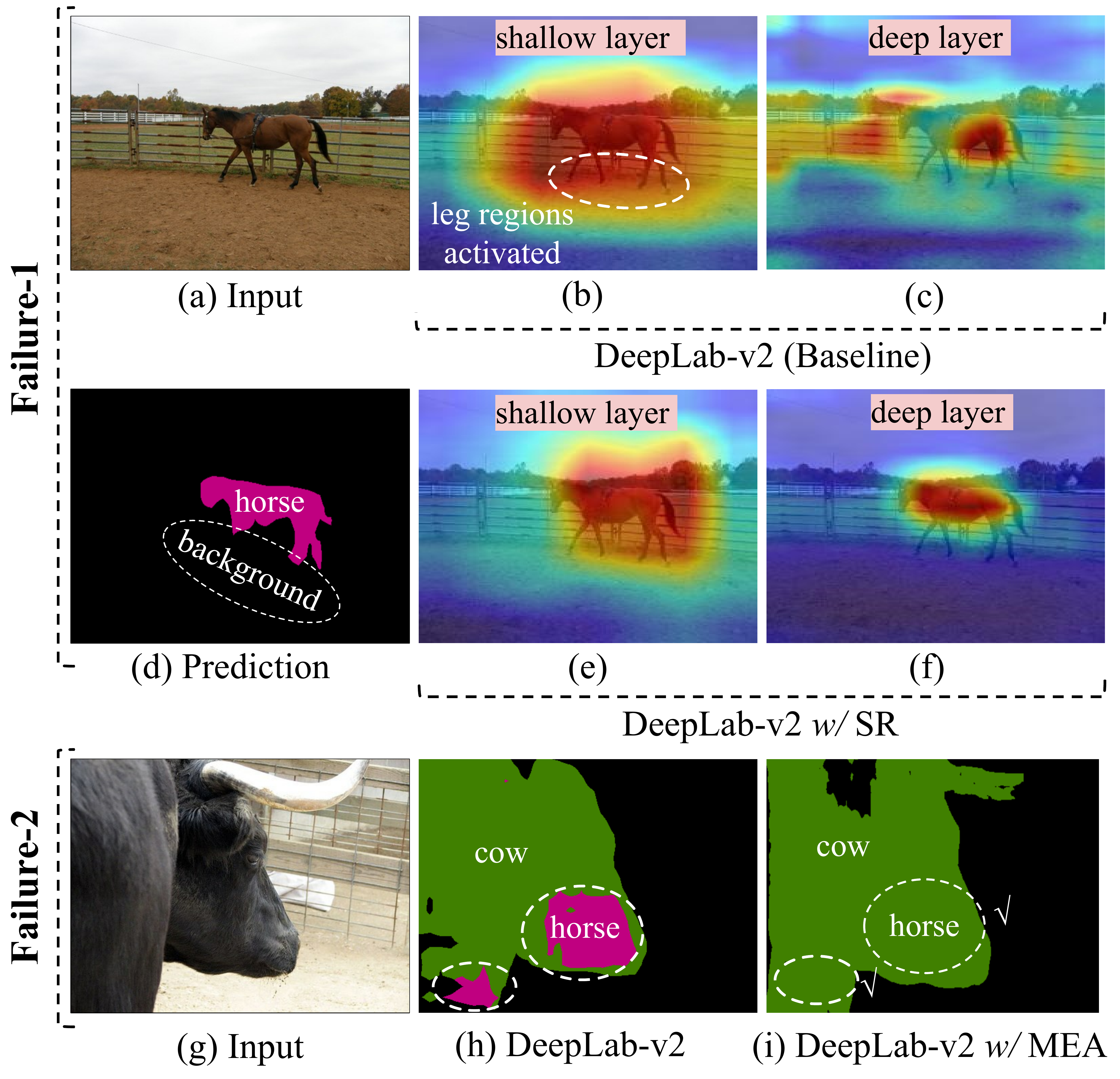}
\caption{
Our inspections on two failure cases of  the baseline method DeepLab-v2~\cite{chen2017deeplab} in (d) and (h), for which (a) and (g) are input images, respectively.
The case in (d) has the flaw of missing ``horse legs''. The visualization of shallow-layer features in (b) shows that the model indeed attends to ``horse legs'' but also to background noises. Applying our SR, in (e), reduces noises while maintaining the desirable attention on ``horse legs''. In addition, deep-layer features in (c) and (f) also prove the effectiveness of SR.
The case in (h) confuses between object classes (\ie, ``horse'' and ``cow'').
The revision in (i) is obtained by adding class-level cross-entropy losses (\ie, using our MEA loss on DeepLab-v2).
% and show their comparison before and after applying our SR.
%
% The columns marked with ``our inspection'' illustrate our motivation, \eg, visualizing activated regions on the shallow-/deep-layer feature maps, and adding classification loss in training.
% It is clear that shallow-layer preserves the ``horse'' legs which are missing in the prediction, and deep-layer captures the accurate semantic regions of the ``horse''. After adding our ``SR'', both the shallow and deep features can fit the ``horse'' more perfectly than before.
% Moreover, the classification loss reduces the class confusion (\ie, ``horse'' and ``cow'').
%First two columns are the examples of failure cases using baseline models~\cite{chen2017deeplab}. The last column illustrates what we get from specific inspection methods, \eg, visualizing activated regions on the shallow-layer feature map and adding classification loss in training. It is clear that shallow-layer preserves the ``horse'' legs which are missing in the prediction, and classification loss reduces the class confusion (\ie, ``horse'' and ``cow'').
}
\vspace{-4mm}
\label{fig1}
\end{figure}

Semantic Segmentation (SS) aims to label each image pixel with its corresponding semantic class~\cite{long2015fully}. It is an indispensable computer vision building block in the scene understanding systems, \eg, autonomous driving~\cite{treml2016speeding} and medical imaging~\cite{havaei2017brain}.
Thanks to the development of deep convolutional neural networks~\cite{he2016deep,WangSCJDZLMTWLX19} and the labour input in pixel-level annotations~\cite{cordts2016cityscapes,mottaghi2014role}, the research for SS has experienced a great progress, \eg, top-performing models can segment about $85\%$ objects in the complex natural scenes~\cite{liu2020efficientfcn}.
% street scenes~\cite{YuanCW19}.
%
Intriguingly, we particularly seek reasons for the $15\%$ failure cases which generally fall into two categories: 1) missing small objects or minor object parts, \eg, the ``horse legs'' in Figure~\ref{fig1} (d), and 2) mislabeling minor parts of large objects as wrong classes, \eg, a part of ``cow'' is wrongly marked as ``horse'' in Figure~\ref{fig1} (h).

%------------------------------------------------------------------------------------
To this end, we present our empirical inspections
in Figure~\ref{fig1}.
We find in Figure~\ref{fig1} (b) that the minor object parts are clearly visible in the shallow-layers of the model, making use of which could address the issue in Failure-1, as in Figure~\ref{fig1} (e).
% {\color{blue}Besides, using deep-layers is also beneficial for shallow-layers in reducing noises.}
In the literature, there is indeed a popular solution --- reshaping and then combining feature maps from different layers for prediction, \eg, Hourglass~\cite{newell2016stacked}, SegNet~\cite{badrinarayanan2017segnet}, U-Net~\cite{ronneberger2015u} and HRNet~\cite{WangSCJDZLMTWLX19}.
Such implementation typically relies on one of the three specific operations: pixel-wise addition~\cite{badrinarayanan2017segnet,Liu_2021_CVPR,ronneberger2015u}, map-wise concatenation~\cite{chen2017deeplab,WangSCJDZLMTWLX19,zhao2017pyramid} and pixel-wise transformation~\cite{wang2018non,zhang2020feature,zhu2019deformable}. However, they are expensive for deep backbones, \eg, SPGNet~\cite{cheng2019spgnet} and HRNet~\cite{WangSCJDZLMTWLX19} take $2.04 \times$ and $1.87 \times$ of model parameters more than their baselines, respectively.
%
% In this paper, we propose a new operation that can incorporate the knowledge learned at different layers but introduce little overhead for training and none at all for testing.

Besides, what we observe from the ``cow'' example in Figure~\ref{fig1} is that penalizing the model with the image-level classification loss can mitigate Failure-2~\cite{zhang2019co,zhang2018exfuse}: pixel-level confusion between the foreground objects (\ie, the ``cow'') and the local pixel cues (\ie, the ``horse'' part). The intuitive reason is that this loss penalizes the prediction of an unseen class-level context (\ie, ``cow'' can not have a ``horse'' mouth). In this paper, we make use of this intuition and introduce a formal loss function to enhance the contextual encoding for each image.

\begin{figure}[t]
\centering
\includegraphics[width=.42\textwidth]{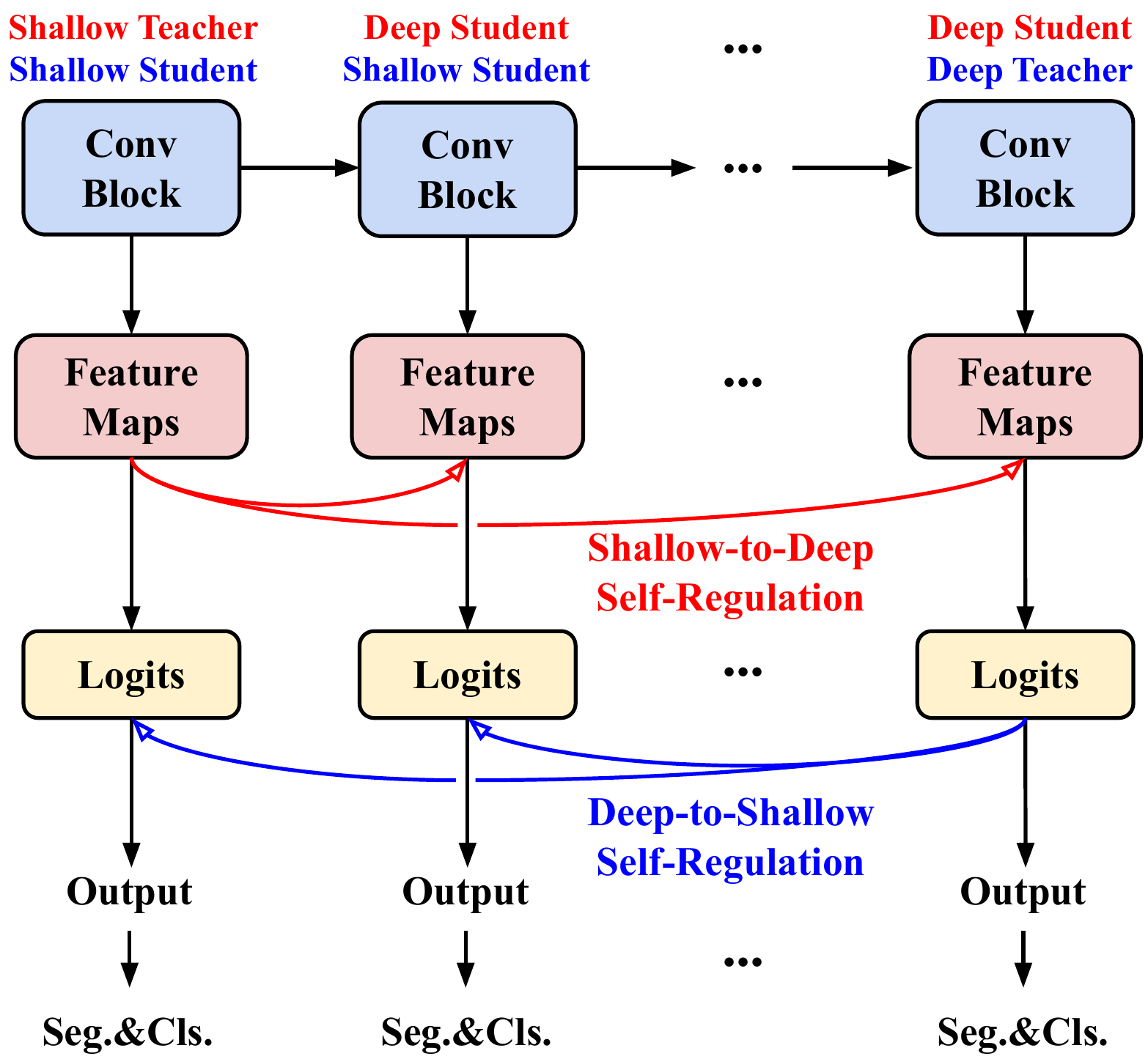}
\caption{An illustrative example of implementing our proposed self-regulation (SR) to a plain Conv network.
It is first converted to a multiple-exit architecture (MEA) where ``Seg.$\&$Cls.'' are a pair of classifier and segmenter for each layer.
Shallow-to-deep SR means shallow-layer features regulate deep layers. Deep-to-shallow SR means deep-layer predictions regulate shallow layers.}
\label{fig2}
\vspace{-4mm}
\end{figure}

Now, we present our overall approach, called \emph{Self-Regulation} (SR), which has three major advantages: 1) generic to be implemented in any deep backbones, 2) without using any auxiliary data or supervision, and 3) with little overhead for training and none for testing. %
As shown in Figure~\ref{fig2}, given a backbone network, we first add a pair of classifier and segmenter at each layer (\eg, head networks for multi-label classification and semantic segmentation), and feed them with the corresponding feature maps.
Then, we apply the proposed Self-Regulation for the features and logits by introducing the following three operations:
\textbf{\uppercase\expandafter{\romannumeral1}}) Regulating the predictions of each pair with ground-truth labels, \ie, the image-level classification labels and the pixel-level mask labels, respectively;
\textbf{\uppercase\expandafter{\romannumeral2}}) Using the feature map of the \emph{shallowest} layer --- Shallow Teacher --- to regulate \emph{all the subsequent} deeper-layer feature maps --- Deep Students;
\textbf{\uppercase\expandafter{\romannumeral3}}) Using the prediction of the \emph{deepest} layer classification logits --- Deep Teacher --- to regulate \emph{all the previous} shallow-layer classification logits --- Shallow Students.
Here is our punchline: for pixel-level feature maps, we use one {\color{red}Shallow Teacher} to teach many {\color{red}Deep Students}; for image-level classification logits, we use one {\color{blue}Deep Teacher} to teach many {\color{blue}Shallow Students}.

%------------------------------------------------------------------------------------
Operation~\textbf{\uppercase\expandafter{\romannumeral1}} is the standard supervised training that spreads over all the layers.
This computation overhead (\eg, multiple-exit network~\cite{phuong2019distillation}) allows the ground-truth to supervise each layer at its earliest convenience.
Operation~\textbf{\uppercase\expandafter{\romannumeral2}} and Operation~\textbf{\uppercase\expandafter{\romannumeral3}} aim to balance the trade-off among these supervisions. To implement them, we introduce a self-regulation method inspired by knowledge distillation (KD)~\cite{hinton2015distilling,liu2020structured, passalis2020heterogeneous}.
Different from the conventional KD that needs to train a teacher model in prior, our ``distillation'' occurs in the same model during training.
The shallowest layer retains the most details, so its segmentation behavior (\eg, encoded in its feature maps and then decoded by the segmentation logits) can teach the subsequent deeper layers not to forget the details to avoid feature underuse (\eg, Failure-1 in Figure~\ref{fig1}); while the deepest layer retain the highest-level contextual semantics, so its classification behavior (\ie, class logits) can teach the previous shallower layers not to ignore the contexts to avoid semantic underuse (\eg, Failure-2 in Figure~\ref{fig1}).

Note that Operation~\textbf{\uppercase\expandafter{\romannumeral1}} at each layer also protects any overuse. Specifically, if Operation~\textbf{\uppercase\expandafter{\romannumeral2}} unnecessarily imposes shallow details on deep contexts, the image-level classification logits at deep layers will penalize it; similarly, if Operation~\textbf{\uppercase\expandafter{\romannumeral3}} unnecessarily imposes deep contexts on shallow details, the pixel-level segmentation logits at shallow layers will discourage it. Such ``shallow to deep and back'' regulation collaborate with each other to improve the overall segmentation. Our empirical results on different baselines, \eg, DeepLab-v2~\cite{chen2017deeplab}, SegNet~\cite{badrinarayanan2017segnet}, SPGNet~\cite{cheng2019spgnet}, and OCRNet~\cite{YuanCW19}, show that our SR is helpful to balance the use of low-/high-level semantics among different layers.

In summary, our contributions are two-fold: 1) A novel set of SR operations that tackle the two key issues in SS tasks while introducing little overhead for training and none for testing; and 2) Through extensive experiments in both fully-supervised and weakly-supervised SS settings, we validate that the proposed SR operations can be easily plugged-and-play and consistently improve various baseline models by a large margin.
\section{Related Work}\label{sec2}
\myparagraph{Semantic Segmentation (SS).}
%------------------------------------------------------------------------------------
The mainstream SS models are based on Fully Convolutional Networks (FCN)~\cite{long2015fully}.
However, deeper layers of FCN often suffer from two problems: underusing detailed spatial information and insufficient receptive fields.
Recent methods proposed to solve these issues can be classified into two camps: 1) encoder-decoder~\cite{noh2015learning,badrinarayanan2017segnet,cheng2019spgnet} based methods, and 2) atrous/dilated convolution~\cite{yu2015multi} based methods.
In the first camp, there are SegNet~\cite{badrinarayanan2017segnet}, De-convNet~\cite{noh2015learning}, and SPGNet~\cite{cheng2019spgnet}. Their common idea is to progressively aggregate high-resolution feature maps using a learnable decoder. In this way, detailed spatial information is embedded to deeper layers without any refinement on its semantic meaning, \ie, such details may bring background noises to deep layers.
% {\color{blue}{However, for this camp, although the output feature maps have wealthy detailed spatial information, they do not have sufficient semantic information~\cite{yang2018denseaspp,liu2019simple}.}}
%
In the second camp, there are DeepLab-v2~\cite{chen2017deeplab} and PSP~\cite{zhao2017pyramid}, and their main idea is to use atrous/dilated convolutions or pooling operations to enable deeper layers to cover large receptive fields and to capture more contextual information~\cite{chen2017deeplab,zhang2019co,zhang2019dual,zhao2017pyramid}.
% {\color{blue}{
These methods fix the size of output feature maps to be $1/8$ or $1/16$ of the input image, so their output are too compressed to contain sufficient spatial details.
% However, since the spatial size of the output feature maps for the second camp is $1/8$ or $1/16$ of the input, the output feature maps usually lack the detailed spatial information.
In contrast to these two camps, our approach enriches the model with both contextual semantics and spatial details by leveraging optimization-based interactions between deep and shallow layers in FCN.

\myparagraph{Multiple-Exit Architecture (MEA).}
MEA was firstly proposed in~\cite{huang2017multi} for training image recognition models, aiming to improve the inference speed of the models~\cite{phuong2019distillation}.
Recently, MEA has been also successfully applied to boost the model performance~\cite{luan2019msd, teerapittayanon2016branchynet} in a variety of visual tasks, \eg, image classification~\cite{duggal2020elf, zhang2018graph}, object detection~\cite{yang2020mutualnet}, and semantic segmentation~\cite{zhang2018context, zhang2018exfuse}.
Our differences with SS-related works~\cite{zhang2018context, zhang2018exfuse}
% include two points. 1) Our aim is to mitigate the pixel-level confusion among similar classes. 2)
lie in that our implementation of MAE integrates two visual tasks, \ie, multi-label classification and semantic segmentation, in a single deep model.

\myparagraph{Knowledge Distillation (KD).}
KD uses a pre-trained teacher model to produce soft labels as ground truth to train a student model~\cite{hinton2015distilling}.
It was originally introduced to compress a big network (teacher) to a small network (student), \ie, network compression~\cite{hui2018fast,tung2019similarity}.
It computes a distillation loss between teacher and student, for which the implementation methods can be logits-based~\cite{jiang2019knowledge,phuong2019distillation}, feature-based~\cite{liu2020deep,xu2020feature}, and hybrid~\cite{li2017mimicking,xu2020knowledge}.
In recent works, a self learning version called self KD, \ie, teacher and student networks have the same architecture, achieved impressive results in several visual tasks, such as image classification~\cite{urban2016deep, xu2020feature}, object detection~\cite{li2017mimicking, liu2020deep}, and pedestrian re-identification~\cite{chen2017darkrank, fan2019spherereid}.
Our approach is based on layer-wise self KD~\cite{zhang2019your,xu2019data}. We introduce a novel approach of bi-directional layer-wise teaching $\&$ learning to particularly enhance SS models.
Our model is learned on a single network and our KD supervisions are applied across different layers.
%Existing KD methods for SS can be divided into three categories: pixel-wise distillation~\cite{liu2019structured}, local patch-based distillation~\cite{xie2018improving}, and holistically-structured distillation~\cite{liu2020structured}.
Compared to the existing KD methods, our approach is different in two aspects. First, it is self-regulated and does not require any pre-trained model as teacher. Second, it consists of two distillation directions each for a specific purpose of learning, \ie, the shallow-to-deep regulation is to enrich the details in the deeper feature maps and the deep-to-shallow one is to force shallow layers to pay more attention to contexts.
%------------------------------------------------------------------------------------
\section{Preliminaries}
\label{sec3}
Our approach essentially encourages layer-wise interactions within a SS model.
For the convenience of comparison, we make a brief review of closely related methods (which also focus on layer-wise interactions).
We divide these methods into three types according to their operations.
1) \textbf{Type 1}: Pixel-wise Addition,
2) \textbf{Type 2}: Map-wise Concatenation, and 3) \textbf{Type 3}: Pixel-wise Transformation.

\noindent\textbf{Type 1} sums up feature maps element-wisely, which are from two layers. It is an intuitive feature interaction method, and has been widely deployed in the U-Shape networks, \eg, UNet~\cite{ronneberger2015u}, SegNet~\cite{badrinarayanan2017segnet}, and SPGNet~\cite{cheng2019spgnet}.
Please note that before the addition, one set of feature maps must be resized to match the size of the other sets as in~\cite{he2016deep,lin2017feature,noh2015learning,passalis2020heterogeneous}.
Popular resizing methods include bi-linear interpolation~\cite{smith1981bilinear} (or deconvolution~\cite{noh2015learning}) for up-sampling and stride convolution (or pooling) for down-sampling.

\noindent\textbf{Type 2} concatenates feature maps (before which one set of feature maps are resized) along the channel dimension, \eg, ASPP~\cite{chen2017deeplab}, PSP~\cite{zhao2017pyramid}, HRNet~\cite{WangSCJDZLMTWLX19}, and Res2Net~\cite{gao2019res2net}.
There is an additional convolution applied to the stacked feature maps for channel size adjustment.
It is indeed through this convolution that the feature maps can interact with each other within local regions.

\noindent\textbf{Type 3} applies the advanced non-local interaction~\cite{wang2018non} densely between every pixel from one feature map and all pixels from another feature map.
This category has been validated of high effectiveness on several tasks, \ie, image classification~\cite{li2020neural,xu2019nonlocal}, object detection~\cite{cao2019gcnet,zhu2019deformable}, and semantic segmentation~\cite{zhang2020feature,zhu2019asymmetric}. However, it is expensive to apply in deep neural networks.

\noindent
\myparagraph{Our difference from the above methods.} We highlight that the above methods are all based on the direct operations on feature maps. In contrast, our self-regulation approach is indirect as it is embodied in the loss function. Specifically, it is realized by applying additional loss terms to the loss function. Therefore, our approach is orthogonal to them. Our experiments show that it can play as an easy and cheap plug-and-play to them, while bringing consistent and significant performance boosts.

% either fail to take into account the representative differences among different layers and ignore the thought of ``learning'' interactions (\ie, the first and the second category), or the computation overhead is extremely large and bring expensive computational cost (\ie, the third category). In our work, we propose an effective yet efficient module, termed self-regulation, which is based on optimization with almost no computation overhead. Besides, self-regulation has high flexibility, which can be implemented on different segmentation models.
%------------------------------------------------------------------------------------
\section{Self-Regulation (SR)}
\label{sec4}
\begin{figure*}[t]
\centering
\includegraphics[width=.95\textwidth]{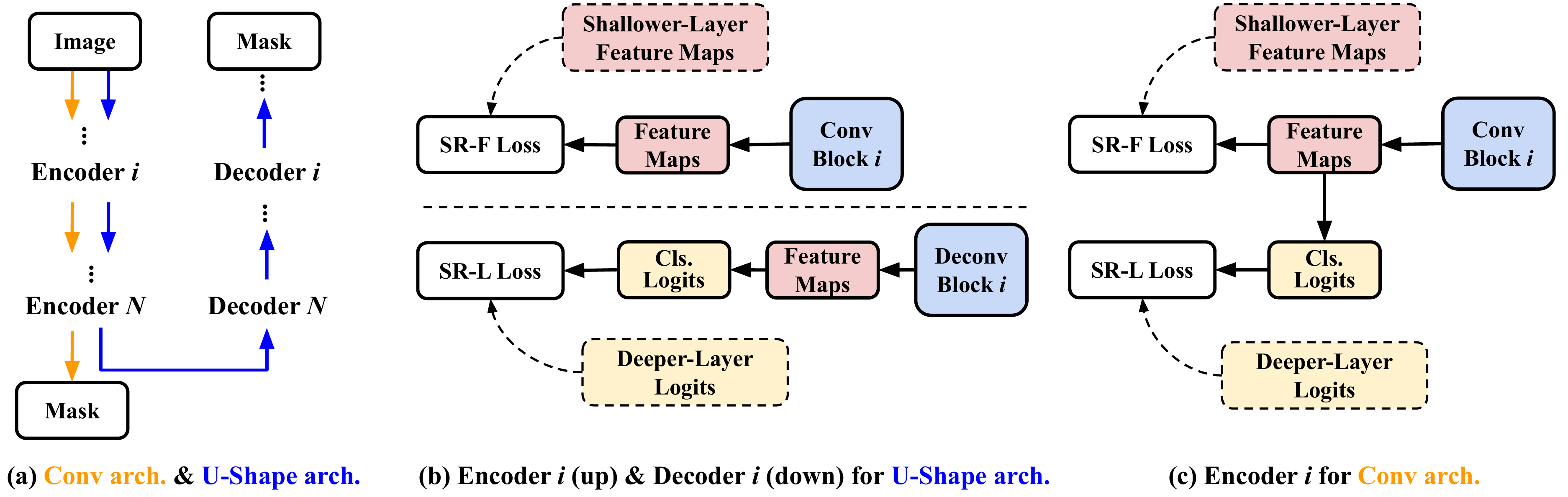}
\caption{Two example network architectures (arch) in (a) and their respective ways of applying SR in (b) and (c).
``Encoder (Decoder) $i$'' denotes the $i$-th Conv (Deconv) block and its related computation in the Encoder (Decoder) network of U-Shape arch, or in the Conv arch (where there is no Decoder).
Both architectures have $N$ Conv blocks, and U-Shape arch additionally contains $N$ Deconv blocks. ``Cls. Logits'' denotes the classification Logits.}
% Please note that in our baseline }
% ``C'' denotes the dataset class size.}
\label{fig3}
\vspace{-4mm}
\end{figure*}

Our proposed SR applies three sets (kinds) of losses for training semantic segmentation models: 1) Multi-Exit Architecture losses~\cite{huang2017multi,phuong2019distillation} (\textbf{MEA})---image-level multi-label classification loss and pixel-level segmentation loss for every block (layer); 2) Self-Regulation losses using Feature maps (\textbf{SR-F})---distillation losses between the features output by Teacher and Student blocks; and 3) Self-Regulation losses using classification Logits (\textbf{SR-L})--- distillation losses between the classification logits of Teacher and Student blocks. For the details of MEA losses, please kindly refer to the paragraph headed with \textbf{MEA} in  Section~\ref{sec2}.
For SR-F and SR-L losses, we demonstrate their computation pipelines in Figure~\ref{fig3} based on two popular semantic segmentation backbones: a fully convolutional architecture (Conv arch)~\cite{long2015fully} and a U-Shape Encoder-Decoder architecture (U-Shape arch)~\cite{cheng2019spgnet}.

In the standard U-Shape arch, each Encoder has a sequence of Conv blocks and every Decoder has the same number of Deconv blocks.
In its Encoder, feature maps output by the shallowest block are taken as the ``ground truth features'' to teach deeper blocks by computing SR-F losses.
For its Decoder (or the Encoder of Conv arch), the classification logits output by the deepest block are taken as the ``ground truth logits'' to teach shallower blocks by computing SR-L losses.
When back-propagating these losses, the model is optimized to capture more pixel-level (taught by the shallowest block) as well as semantic-level (taught by the deepest block) information in each layer, thus derives a more powerful image representation.

% I comment on the following parag for saving space, if you have space, you can release it:
% Below we introduce the detailed formulations of SR-F and SR-L losses, and explain how to implement them on two prevalent SS frameworks~\cite{cheng2019spgnet,WangSCJDZLMTWLX19}.
\subsection{SR-F Loss: Shallow to Deep}
\label{sec4_1}
%%
% We clarify the definitions and notations for the Teacher and the Student in our MEA.
As illustrated in Figure~\ref{fig3}, each Conv block or Deconv block can be regarded as an individual feature extractor.
% and its features can be used to train a classifier and a segmenter.
%
The ending (the shallowest or the deepest) block can be either teacher or student in different situations, as our self-regulation is bidirectional. The middle blocks are always students.
% Our self-regulation is bidirectional which means the ending (the shallowest or the deepest) block plays a different role (either teacher or student) in each different situations.
%
We denote the transformation functions of Teacher and Student as $\mathcal{T}_{\theta}(\textbf{x})$ and $\mathcal{S}_{\phi}(\textbf{x})$, respectively, where $\theta$ and $\phi$ represent their corresponding network parameters, and $\textbf{x}$ is the input image. For simplicity, we omit $\textbf{x}$ in the following formulations.

% Recall the motivation of SR-F loss is to encourage deep blocks to preserve more detailed information which is well-retained in shallow blocks.
% In this paper, we simply deploy the shallowest (\ie, the first) block as the Teacher (called Shallow Teacher) to regulate all the other blocks (called Deep Students), as illustrated in Figure~\ref{fig2}.

SR-F loss aims to encourage deep blocks to preserve the detailed information that is retained in shallow blocks. Our method simply deploys the shallowest (the first) block as the Teacher (called Shallow Teacher) to regulate deep blocks (called Deep Students).
%
% We achieve this regulation by updating the parameters of Deep Students, \ie, backpropagating their feature map distance loss compared to the Shallow Teacher's.
% We call this distance loss: SR-F loss.
Below, we elaborate the computation pipeline of SR-F loss.
%%%
We use a cross-entropy function to measure the distance between Shallow Teacher's feature maps $\mathcal{\textbf{T}}^{[1]}_{\theta}$ to the feature maps of the $i$-th Deep Student (\ie, the $i$-th block) $\mathcal{\textbf{S}}^{[i]}_{\phi}$, and obtain the SR-F loss as follows,
\vspace{-2mm}
\begin{equation}
  \mathcal{L}_\textrm{{ce}}\big(\mathcal{\textbf{T}}^{[1]}_{\theta},\mathcal{\textbf{S}}^{[i]}_{\phi}\big) =
  -\frac{1}{M}\sum_{j=1}^M {\sigma(\textbf{t}_j)\textrm{log}\left(\sigma(\textbf{s}_j)\right)},
\label{eq1}
\end{equation}
where $i \in [2, N]$, $N$ denotes the number of Conv blocks in a Conv arch (or a U-Shape arch);
$\textbf{t}_j$ is the vector on the $j$-th pixel position in $\mathcal{\textbf{T}}^{[1]}_{\theta}$, and its dimension is the same as the channel size; $\textbf{s}_j$ is the corresponding vector in $\mathcal{\textbf{S}}^{[i]}_{\phi}$;
% abs($\cdot$) produces the absolute value;
$\sigma$ denotes the softmax normalization along channels.
and $M$ is the spatial size ($=$ width $\times$ height) of the feature map.

In general KD~\cite{hinton2015distilling}, an advanced technique to improve efficiency is to leverage the temperature scaling~\cite{guo2017calibration}. If applying this technique in our case, we can rewrite Eq.~\ref{eq1} as:
\vspace{-3mm}
\begin{equation}
  \mathcal{L}^{\tau}_\textrm{{ce}}\big(\mathcal{\textbf{T}}^{[1]}_{\theta}, \mathcal{\textbf{S}}^{[i]}_{\phi}\big) =
  -\tau^2\frac{1}{M}\sum_{j=1}^M {\sigma(\textbf{t}_j)^{1/\tau}\textrm{log}\left(\sigma(\textbf{s}_j)^{1/\tau} \right)
  },
\label{eq2}
\end{equation}
where we note that the value of $\textbf{t}_j^{1/\tau}$ (or $\textbf{s}_j^{1/\tau}$) is normalized within each feature map.
$\tau \in \mathbb{R}_+$ denotes the temperature and its effect is spatial smoothness, \eg, a larger value of $\tau$ results a greater suppression on the difference between the maximum and minimum pixel values on the feature map.

The Shallow Teacher is ``teaching'' many Deep Students, so the overall SR-F loss $\mathcal{L}_\textrm{{SR-F}}$ can be derived as follows,
\begin{equation}
  \mathcal{L}_\textrm{{SR-F}} = \sum^N_{i=2}\mathcal{L}^{\tau}_\textrm{{ce}}\big(\mathcal{\textbf{T}}^{[1]}_{\theta}, \mathcal{\textbf{S}}^{[i]}_{\phi}\big).
\label{eq3}
\end{equation}

\subsection{SR-L Loss: Deep to Shallow}
\label{sec4_2}

SR-L aims to encourage shallow blocks (layers) to capture more global contextual information, \ie, the shape of a ``horse'', such that their feature can be more robust against background noises.
Our method simply deploys the deepest (\ie, the last) layer as the Teacher (called Deep Teacher) to regulate all shallow layers (called Shallow Students).
We understand that the classification logits contain the high-level semantic information of object classes, so we use those logits of Deep Teacher as ``teaching materials''.

Given Deep Teacher's logits $\mathcal{\textbf{T}}^{[N]}_{\theta}$ and all Shallow Students's logits $\{\mathcal{\textbf{S}}^{[k]}_{\phi}\}_{k=1}^{N-1}$, we just follow methods as in Eq.~\ref{eq1} to Eq.~\ref{eq3}.
\emph{The difference is that we compute cross-entropy  loss between two sets of logits rather than feature maps}.
Therefore, we obtain the SR-L loss as follows,
\vspace{-2mm}
\begin{equation}
  \mathcal{L}_\textrm{{SR-L}} = \sum^{N-1}_{k=1}\mathcal{L}^{\tau}_\textrm{{ce}}\big(\mathcal{\textbf{T}}^{[N]}_{\theta}, \mathcal{\textbf{S}}^{[k]}_{\phi}\big).
\label{eq4}
\end{equation}

\subsection{Overall Loss Function}
\label{sec4_3}
%%
% So far we have introduced two important SR loss terms. We then incorporate them
Below we incorporate the introduced SR losses with the layer-wise loss terms for multi-label classification as well as semantic segmentation (computed from multiple pairs of classifier and segmenter as illustrated in Figure~\ref{fig2}), and derive the overall objective function as follows,
\vspace{-2mm}
\begin{equation}
    \mathcal{L}_\textrm{{SR}} =
    \underbrace{\lambda_1\mathcal{L}_\textrm{{SR-cls}} + \lambda_2\mathcal{L}_\textrm{{SR-seg}}}_{\textrm{MEA~loss}} + \lambda_3\mathcal{L}_\textrm{{SR-F}} + \mathcal{L}_\textrm{{SR-L}},
\label{eq5}
% \vspace{-2mm}
\end{equation}
where $\mathcal{L}_\textrm{{SR-seg}}$ and $\mathcal{L}_\textrm{{SR-cls}}$ are cross-entropy losses using ground truth one-hot labels. \emph{This pair of loss terms are called MEA loss in the following.}
On the top of them, $\mathcal{L}_\textrm{{SR-F}}$ and $\mathcal{L}_\textrm{{SR-L}}$ bring further performance improvement to the model because they encourage the layer in the model to learn from the ``soft knowledge'' acquired by another layer which is beyond the knowledge of one-hot labels.
$\lambda_1$, $\lambda_2$, and $\lambda_3$ denote the weights used to balance these losses.

% {\color{blue}{In the model training process, these losses are dynamically assigned different weights so that the model is optimized.}}

We highlight that our $\mathcal{L}_\textrm{{SR}}$ is a general loss function, which is easy to incorporate into any FCN-based SS model.
In the following, we introduce how to implement $\mathcal{L}_\textrm{{SR}}$ on two representative as well as state-of-the-art SS backbones, \ie, a U-Shape arch and a Conv arch.

\subsection{SR on Specific SS Backbones}
\label{sec4_4}

In Figure~\ref{fig3} (a), we demonstrate an U-Shape arch on the $\textrm{{\color{blue}{Blue}}}$ path and a Conv arch on the \textcolor[RGB]{255,153,0}{Orange} path.
To better elaborate the implementation of SR on these two types of backbones, we take the state-of-the-art SPGNet~\cite{cheng2019spgnet} and HRNet~\cite{WangSCJDZLMTWLX19} as examples, respectively.

\myparagraph{SPGNet.}
It is a two-stage U-Shape arch (double-sized compared to the U-Shape arch in Figure~\ref{fig3} (a)), each stage is a standard U-Shape arch consisted of $5$ Conv and $5$ Deconv residual blocks~\cite{cheng2019spgnet}.
%Therefore, $N=10$ in SPGNet.
In Figure~\ref{fig3} (b), we illustrate our method of applying SR-F and SF-L losses to an example pair of Conv and Deconv blocks.

\myparagraph{In the $i$-th Conv block} ($1<i\le N$), a $3 \times 3$ convolution is first deployed to reduce the channel size to $256$.
Another $3 \times 3$ convolutional layer followed by a batch normalization layer is then used to generate the feature maps $\mathcal{\textbf{S}}^{[i]}_{\phi}$ which have the same channel size
%\footnote{For a fast computation, feature maps from different layers (blocks) are reshaped into $1/8$ spatial size of the input image.}
with the feature maps $\mathcal{\textbf{T}}^{[1]}_{\theta}$ of Shallow Teacher.
Next, we compute the $i$-th SR-F loss $\mathcal{L}^{[i]}_\textrm{{SR-F}}$ following Eq.~\ref{eq2}.
Finally, we can derive the overall SR-F loss, \ie, $\mathcal{L}_\textrm{{SR-F}}$ by summing up the SR-F losses of all Deep Students $\big\{\mathcal{\textbf{S}}^{[i]}_{\phi}\big\}^N_{i=2}$ as in Eq.~\ref{eq3}.

\myparagraph{In the $i$-th Deconv block} ($1\le i<N$), we follow the same method to generate feature maps, \ie, through two convolutional layers and one batch normalization layer.
Then, we feed these maps into a pair of classifier and segmenter.
We use the resulted predictions to respectively compare to image-level and pixel-level one-hot labels, yielding two standard cross-entropy losses, \ie, $\mathcal{L}^{[i]}_\textrm{{SR-cls}}$ and $\mathcal{L}^{[i]}_\textrm{{SR-seg}}$.
Computing these losses through all Deconv blocks, we can get the overall MEA loss (consisting of $\mathcal{L}_\textrm{{SR-cls}}$ and $\mathcal{L}_\textrm{{SR-seg}}$).

Besides, we denote the classification logits of Deep Teacher as $\mathcal{\textbf{T}}^{[N]}_{\theta}$, and use them to regulate each Shallow Student $\mathcal{\textbf{S}}^{[i]}_{\phi}$, \ie, to compute the SF-L loss $\mathcal{L}^{[i]}_\textrm{{SR-F}}$.
When getting all SF-L losses (through all Shallow Students) $\big\{\mathcal{\textbf{S}}^{[i]}_{\phi}\big\}^{N-1}_{i=1}$, we can derive the overall SR-L loss $\mathcal{L}_\textrm{{SR-L}}$ using Eq.~\ref{eq4}.

\myparagraph{HRNet.}
It is a Conv arch consisting of $4$ Conv blocks, \ie, $N$=$4$~\cite{WangSCJDZLMTWLX19}. Its block 1 produces the highest resolution feature maps. These feature maps are fed into two paths: path 1 goes to block 2 to continue producing high-resolution feature maps; and path 2 is down-sampling first via a stride $3 \times 3$ convolution and then goes to block 2 to produce low-resolution feature maps. Repeating these two paths on block 3 and block 4 yields four outputs with different resolutions.

We take block $i$ as an example and illustrate the computation flow in Figure~\ref{fig3} (c). First, we apply a $3 \times 3$ convolution and a batch normalization layer to reduce the channel size of feature maps to be $256$.
Then, we resize those maps to be $1/8$ spatial size of the input image, and apply another $3 \times 3$ convolution to
produce the final feature maps (named as ``Feature Maps'' in Figure~\ref{fig3} (c)) of block $i$.
We take the highest-resolution feature (after block 1) as Shallow Teacher $\mathcal{\textbf{T}}^{[1]}_{\theta}$. The SR-F loss is thus the summation of its distances to all Deep Students $\big\{\mathcal{\textbf{S}}^{[j]}_{\phi}\big\}^{N}_{k=2}$ as in Eq.~\ref{eq3}.
We take the classification logits of block 4 as Deep Teacher $\mathcal{\textbf{T}}^{[N]}_{\theta}$. The SR-L loss is thus the summation of its distances to all Shallow Students $\big\{\mathcal{\textbf{S}}^{[j]}_{\phi}\big\}^{N-1}_{k=1}$ as in Eq.~\ref{eq4}.
The MEA loss consisting of $\mathcal{L}_\textrm{{SR-seg}}$ and $\mathcal{L}_\textrm{{SR-cls}}$ can be derived in the same way as in the U-Shape arch. When using these losses as in Eq.~\ref{eq5}, we apply different combination weights and will show the details in the experiment section.
% , we use these multiple losses to train SS models.
%------------------------------------------------------------------------------------
%------------------------------------------------------------------------------------
\section{Experiments}
%------------------------------------------------------------------------------------
Our SR was evaluated on two SS tasks: \ie, weakly-supervised SS (WSSS) with image-level class labels, and fully-supervised SS (FSSS) with pixel-level class labels.
%
% For both tasks, we reported the quantitative results of an ablation study and extensive comparisons to the state-of-the-art (SOTA) methods, and also showed qualitative visualizations.
% Before these, we first introduced the common and specific settings for two tasks.
%%

\myparagraph{Common Settings.}
All experiments were implemented on PyTorch~\cite{paszke2019pytorch}.
All models were pre-trained on ImageNet~\cite{deng2009imagenet}, and then fine-tuned on the \emph{training} sets of SS as in~\cite{chen2017deeplab,YuanCW19}.
For evaluation\footnote{Please kindly refer to more details about hyperparameters and evaluation results in supplementary materials.}, the mean Intersection of Union (mIoU) was used as the primary evaluation metric. In addition, the number of model parameters and FLOPs were used as efficiency metrics.
Following~\cite{zhang2018context,zhang2019co}, the combination weights of classification loss and segmentation loss were respectively set to $0.2$ and $0.8$ in MEA. The weights of SR-L and SR-F were set equal, \ie, $\lambda_1 = 0.2, \lambda_2 = 0.8, \lambda_3 = 1$. The results of using other weights are given in the supplementary materials.
For the hyperparameter $\tau$ (temperature), we applied an adaptive annealing scheme, \ie, initializing $\tau$ as $1$ and multiplying it by a factor of $1.05$ every time when the difference between the minimum and maximum values (on any feature map in the minibatch) exceeds $0.5$.

\subsection{WSSS Settings}
\myparagraph{Datasets.}
Our WSSS experiments were carried out on two widely-used benchmarks: PASCAL VOC 2012 (PC)~\cite{everingham2010pascal} and MS-COCO 2014 (MC)~\cite{lin2014microsoft}.
PC dataset consists of $21$ classes ($20$ for objects and $1$ for background) and splits $1,464$ images for \emph{training}, $1,449$ for \emph{val}, and $1,456$ for \emph{test}.
Following related works~\cite{ahn2019weakly, huang2018weakly, kolesnikov2016seed, zhang2020causal}, we used an enlarged \emph{training} set including $10,582$ images.
MC dataset consists of $81$ classes ($80$ for objects and $1$ for background) with $80k$ and $40k$ images respectively for \emph{training} and \emph{val}.
In the training phase of WSSS, \textbf{only the image-level class labels were used as ground truth}.
For data augmentation, we followed the same strategy as in~\cite{zhang2018context}, including gaussian blur, color augmentation, random horizontal flip, randomly rotate (from $-10^{\circ}$ to $+10^{\circ}$), and random scaling (using the scaling rates between $0.5 \times$ and $2 \times$).

\myparagraph{Baseline Models.}
The WSSS framework includes two steps: pseudo-mask generation and semantic segmentation training with pseudo-masks. For pseudo-mask generation, we deployed two popular ones, namely IRNet~\cite{ahn2019weakly} and SEAM~\cite{wang2020self}, and the state-of-the-art (SOTA) CONTA~\cite{zhang2020causal}.
For semantic segmentation, we deployed DeepLab-v2~\cite{chen2017deeplab} with ResNet-38~\cite{wu2019wider}, SegNet~\cite{badrinarayanan2017segnet} with ResNet-101~\cite{he2016deep} and the SOTA SPGNet~\cite{cheng2019spgnet} with ResNet-50~\cite{he2016deep}, respectively.
Baseline models are trained on the conventional arch and using only the cross-entropy loss.
Baseline+SR models are ours which apply our proposed SR loss terms (see Eq.~\ref{eq5}) and train the model on the MEA arch (see Figure~\ref{fig2}).

\myparagraph{Training Details.}
Our major settings followed close related works~\cite{ahn2019weakly, cheng2019spgnet, wang2020self, zhang2020causal}.
The input images were cropped into fixed sizes as $512\times 512$ and $448\times 448$ for PC and MC, respectively. Zero padding was used if needed.
The mini-batch SGD momentum optimizer was used to train all SS models with batch size as $16$, momentum as $0.9$ and weight decay as $0.0001$.
The initial learning rates (LR) were $0.01$ and $0.04$ for PC and MC datasets, respectively.
A ``poly'' schedule for LR was deployed, \ie, updating LR as $(1-\frac{iter}{iter_{max}})^{0.9}$.
All our implemented models were trained for $80$ epochs on PC and $50$ epochs on MC.

\subsection{FSSS Settings}
\myparagraph{Datasets.}
Our FSSS experiments were carried out on two challenging benchmarks: Cityscapes (CS)~\cite{cordts2016cityscapes}, and PASCAL-Context (PAC)~\cite{mottaghi2014role}.
CS dataset consists of $19$ classes with $2,975$ images for \emph{training}, $500$ for \emph{val} and $1,525$ for \emph{test}. Please note that we only used these finely annotated images, although this dataset offers $20,000$ coarsely annotated images.
For the PAC dataset, we followed~\cite{zhang2018context, YuanCW19, he2019adaptive} and used the most popular version consisted of $60$ classes (including the background class). There are $4,998$ and $5,015$ images for \emph{training} and \emph{test}, respectively.
For data augmentation, we followed~\cite{YuanCW19, zhang2019co, zhang2020feature} to use random horizontal flipping, random scaling (using the the scaling rates between $0.5 \times$ and $2 \times$), and random brightness jittering (in the range from $-10^{\circ}$ to $+10^{\circ}$).

\myparagraph{Baseline Models.}
We implemented our SR onto a popular method DBES~\cite{li2020improving} with ResNet-101~\cite{he2016deep}, and the SOTA method OCRNet~\cite{YuanCW19} with HRNetV2-W48~\cite{WangSCJDZLMTWLX19}. Please note that HRNetV2-W48 is a backbone network.

\myparagraph{Training Details.}
Our major settings were the same as those in baseline methods~\cite{WangSCJDZLMTWLX19,YuanCW19}.
The input images were cropped into $969 \times 969$ and $520 \times 520$ on CS and PAC datasets, respectively.
The mini-batch SGD momentum optimizer was used in the training phase with batch size as $8$ and momentum as $0.9$.
The initial LRs were set to $0.01$ and $0.001$ for CS and PAC, respectively. The same ``poly'' schedule was deployed.
The L2 regularization term weights were set to $0.0005$ and $0.0001$ for CS and PAC, respectively.
All our implemented models were trained from $580$ epochs on CS and $100$ epochs on PAC.

%------------------------------------------------------------------------------------
\begin{table}[t]
\small
\begin{center}
\renewcommand\arraystretch{1.1}
\setlength{\tabcolsep}{6pt}{
\begin{tabular}{ c  c  c | c | c | c}
MEA & SR-F & SR-L &PC & MC & CS \\
\hline \hline
\xmark & \xmark & \xmark & 67.1 & 33.6 & 80.7$^\flat$  \\
\cmark & \xmark & \xmark & 67.7$_{\color{red}{ +0.6}}$ & 34.0$_{\color{red}{ +0.4}}$ & 81.1$_{\color{red}{ +0.4}}$  \\
\cmark & \xmark & \cmark & 68.1$_{\color{red}{ +1.0}}$ & 34.3$_{\color{red}{ +0.7}}$ & 81.6$_{\color{red}{ +0.9}}$  \\
\cmark & \cmark & \xmark & 68.2$_{\color{red}{ +1.1}}$ & 34.3$_{\color{red}{ +0.7}}$ & 81.7$_{\color{red}{ +1.0}}$  \\
\cmark & \cmark & \cmark & 68.5$_{\color{red}{ +1.4}}$ & 34.5$_{\color{red}{ +0.9}}$ & 82.1$_{\color{red}{ +1.4}}$   \\
\end{tabular}}
\vspace{1mm}
\caption{Ablation study results (mIoU, \%) on the \emph{val} sets of three datasets: PASCAL VOC 2012 (PC)~\cite{everingham2010pascal}, MS-COCO 2014 (MC)~\cite{lin2014microsoft} with image-level class labels (WSSS), and Cityscapes (CS) with pixel-level class labels (FSSS). The WSSS baseline is CONTA~\cite{zhang2020causal}+SPGNet~\cite{cheng2019spgnet}, and the FSSS baseline is OCRNet~\cite{YuanCW19}.
Their results are shown in the first row.
``MEA'' indicates the MEA loss.
``SR-F'' and ``SR-L'' indicate the other two losses respectively (see Eq.~\ref{eq5}). $\flat$ means the result is produced by us using public code.}
\label{tab1}
\end{center}
\vspace{-5mm}
\end{table}
%------------------------------------------------------------------------------------
 \begin{table}[t]
\footnotesize
 \begin{center}
 \renewcommand\arraystretch{1.1}
 \setlength{\tabcolsep}{1pt}{
 \begin{tabular}{r|c|c|c|c|c}
 ~ & Baseline & +MEA & +(MEA,SR-L) & +(MEA,SR-F) & +SR\\
 \hline \hline
 \multicolumn{6}{c}{WSSS; SPGNet~\cite{cheng2019spgnet} as Baseline}\\
 \hline
 Params. & 55.6\textbf{M} & 56.7$_{\color{red}{ +1.1}}$\textbf{M} & 56.7$_{\color{red}{ +0.0}}$\textbf{M} & 56.9$_{\color{red}{ +0.2}}$\textbf{M} & 56.9\textbf{M} \\
 FLOPs & 467.6\textbf{B} & 467.9$_{\color{red}{ +0.3}}$\textbf{B} & 467.9$_{\color{red}{ +0.0}}$\textbf{B} & 467.9$_{\color{red}{ +0.0}}$\textbf{B} & 467.9\textbf{B} \\
 \hline
  \multicolumn{6}{c}{FSSS; OCRNet~\cite{YuanCW19} as Baseline}\\
 \hline
 Params. & 76.4\textbf{M} & 78.9$_{\color{red}{ +2.5}}$\textbf{M} & 78.9$_{\color{red}{ +0.0}}$\textbf{M} & 79.5$_{\color{red}{ +0.6}}$\textbf{M} & 79.5\textbf{M} \\
 FLOPs & 1,087.3\textbf{G} & 1,095.5$_{\color{red}{ +8.2}}$\textbf{G} & 1,095.5$_{\color{red}{ +0.0}}$\textbf{G} & 1,095.5$_{\color{red}{ +0.0}}$\textbf{G} & 1,095.5\textbf{G} \\
\end{tabular}}
\vspace{1mm}
\caption{Model efficiency analysis for different components of SR. ``+SR'' means including all terms (MEA, SR-L, and SR-F).}
\label{tab2}
\end{center}
\vspace{-6mm}
\end{table}
%------------------------------------------------------------------------------------

\subsection{Results and Analyses}
\myparagraph{Ablation Study.}
We conducted the ablation study on the val sets of PC, MC and CS datasets, and reported the results in Table~\ref{tab1}. ``MEA'' denotes using the MEA loss consisting of $\mathcal{L}_\textrm{{SR-seg}}$ and $\mathcal{L}_\textrm{{SR-cls}}$.
We used the SOTA WSSS method, namely CONTA~\cite{zhang2020causal}+SPGNet~\cite{cheng2019spgnet}, as the baseline here and show its results in row 1.
Comparing row 5 to row 1, we can see the proposed SR loss brought clear performance gains, \eg, $1.4\%$ on PC dataset.
Intriguingly, comparing row 5 to row 2 (and row 1), we find that distillation-based loss terms ($\mathcal{L}_\textrm{{SR-L}}$ and $\mathcal{L}_\textrm{{SR-F}}$) brought higher improvement margins than using only MEA loss.
As we mentioned under Eq.~\ref{eq5}, this is because our SR-L and SR-F terms encourage each individual layer to learn the ``soft knowledge'' (from a superior layer) which is richer than the ``hard knowledge'' in one-hot labels (used for computing the MEA loss).
This phenomenon is consistent across all datasets.
% We can observe that different components of SR can bring consistent improvements on these two datasets.
%
% For example, compared to the baseline model, MEA can boost $0.6 \%$ and $0.4 \%$ mIoU improvements on PC and MC, respectively.
% Based on this, SR-L and SR-F respectively brings $0.5 \%$ and $0.4 \%$ mIoU improvements on PC. Besides, with the help of MEA, SR-L and SR-F can respectively bring $0.3 \%$ and $0.3 \%$ mIoU improvements on MC as well. In particular, the complete combination of MEA, SR-L, and SR-F results in the best performance, \ie, $68.5 \%$ and $34.5 \%$ mIoU on PC and MC on the \emph{val} sets, which can surpass the baseline model by $1.4 \%$ and $0.9 \%$, respectively.\\
%------------------------------------------------------------------------------------

\myparagraph{Model Efficiency.}
Our SR brings performance improvements without increasing much computational costs.
% {\color{blue}{, and the overhead of SR mainly comes from additional convolutional layers}}.
To validate this, we show the statistics of Params, \ie, the number of network parameters, and FLOPs, \ie, the speed of training, in Table~\ref{tab2}.
% using SR introduced very little computational overhead, regarding both the number of network parameters and the speed of model training.
%
It is clearly shown that SR introduced very little computational overheads for both SS tasks.
For example in WSSS, applying MEA, SR-L and SR-F losses increased $1.1\textbf{M}$, $0.0\textbf{M}$, and $0.2\textbf{M}$ Params, respectively.
Using MEA increased only $0.3\textbf{B}$ FLOPs, and using SR-L and SR-F for $0$.
This overhead is mainly caused by using additional convolutional layers in constructing MEA.
% There are also consistent in the fully-supervised settings.
%% The

\begin{table}[t]
\small
\begin{center}
\renewcommand\arraystretch{1.1}
\setlength{\tabcolsep}{4pt}{
\begin{tabular}{r|c|cc|c}
\multicolumn{1}{r|}{Methods}& Backbone & PC \emph{val} & PC \emph{test}  & MC \emph{val}\\
\hline \hline
% OAA+~\cite{jiang2019integral} & ResNet-101 & 65.2  & 66.4  & -- \\
SCE~\cite{chang2020weakly} & ResNet-101 & 66.1  & 65.9  & -- \\
EME~\cite{fan2020employing} & ResNet-101 & {67.2}  & 66.7  & -- \\
MCS~\cite{sun2020mining} & ResNet-101 & 66.2  & {66.8}  & -- \\
\hline
IRNet~\cite{ahn2019weakly} & Deeplab-v2~\cite{chen2017deeplab} & 63.5  & 64.8  & 32.6\\
IRNet+SR & Deeplab-v2~\cite{chen2017deeplab} & 64.6$_{\color{red}{ +1.1}}$  & 65.8$_{\color{red}{ +1.0}}$  & 33.4$_{\color{red}{ +0.8}}$ \\
 \cdashline{1-5}[0.8pt/2pt]
SEAM~\cite{wang2020self} & Deeplab-v2~\cite{chen2017deeplab} & 64.5  & 65.7  & 31.9 \\
SEAM+SR & Deeplab-v2~\cite{chen2017deeplab} & 65.6$_{\color{red}{ +1.1}}$  & 66.5$_{\color{red}{ +0.8}}$  & 32.6$_{\color{red}{ +0.7}}$ \\
 \cdashline{1-5}[0.8pt/2pt]
CONTA~\cite{zhang2020causal} & Deeplab-v2~\cite{chen2017deeplab} & 66.1  & 66.7  & 33.4 \\
CONTA+SR & Deeplab-v2~\cite{chen2017deeplab} & 66.8$_{\color{red}{ +0.7}}$  & 67.2$_{\color{red}{ +0.5}}$ & 34.0$_{\color{red}{ +0.6}}$\\
 \cdashline{1-5}[0.8pt/2pt]
CONTA~\cite{zhang2020causal}& SegNet~\cite{badrinarayanan2017segnet} & 66.9 & 67.7 & 33.7 \\
CONTA+SR & SegNet~\cite{badrinarayanan2017segnet} & 67.9$_{\color{red}{ +1.0}}$ & 68.4$_{\color{red}{ +0.7}}$ & 34.4$_{\color{red}{ +0.7}}$ \\
 \cdashline{1-5}[0.8pt/2pt]
CONTA~\cite{zhang2020causal} & SPGNet~\cite{cheng2019spgnet} & 67.1  & 67.9 & 33.6  \\
CONTA+SR & SPGNet~\cite{cheng2019spgnet} & {68.5}$_{\color{red}{+1.4}}$  & {69.1}$_{\color{red}{+1.2}}$ & {34.5}$_{\color{red}{ +0.9}}$  \\
\end{tabular}}
\vspace{1mm}
\caption{Comparing to the state-of-the-arts on the \emph{val} and \emph{test} sets of PASCAL VOC 2012 (PC)~\cite{everingham2010pascal}, and the \emph{val} set of MS-COCO (MC) 2014~\cite{lin2014microsoft} using image-level class labels. ``--'' denotes there are no reported results in the original papers. ``+SR'' means applying our SR loss (Eq.~\ref{eq5}) function to train the models.}
\label{tab3}
\end{center}
\vspace{-6mm}
\end{table}

\myparagraph{Comparing to SOTA in WSSS.}
% To compare with SOTA WSSS models, we implemented
% %our approach
% {\color{blue}{
To compare with SOTA methods, we implemented three pseudo-mask generation approaches, \ie, IRNet~\cite{ahn2019weakly}, SEAM~\cite{wang2020self}, and CONTA~\cite{zhang2020causal}, and used three baseline SS models, \ie, DeepLab-v2~\cite{chen2017deeplab}, SegNet~\cite{badrinarayanan2017segnet}, and SPGNet~\cite{cheng2019spgnet}. We plugged our SR in all of them.
We report the results on the \emph{val} and \emph{test} sets of the PC dataset, and the \emph{val} set of the MC dataset in Table~\ref{tab3}.
It is shown that using our SR consistently improved the performance of all implemented methods and achieved the top performance on two datasets.
For example, it boosted DeepLab-v2 on the \emph{val} set of PC by $1.1\%$, $1.1\%$, and $0.7\%$ mIoU improvements when using IRNet, SEAM and CONTA for generating pseudo masks, respectively.
It also boosted SegNet and SPGNet on PC respectively by $1.0\%$ and $1.4\%$ when using CONTA.
On the \emph{test} set of PC, it achieved the best performance which is higher than the SOTA method (CONTA w/ SPGNet) by $1.2\%$ mIoU.
Its superiority is also obvious when comparing its best results to those of the methods in the top block, \eg, it surpassed EME by $1.3\%$ on PC \emph{val} set and MCS by $2.3\%$ on PC \emph{test} set.

\myparagraph{Comparing to SOTA in FSSS.}
In the task of FSSS, SOTA methods include DBES~\cite{li2020improving} on ResNet-101~\cite{he2016deep} (backbone) and OCRNet~\cite{YuanCW19} on HRNet~\cite{WangSCJDZLMTWLX19} (backbone), on both CS and PAC datasets.
We present their original results and also show our results (by plugging SR in these methods) in Table~\ref{tab4} and Table~\ref{tab5} respectively for two datasets.
We can see from both tables that our SR becomes the new state-of-the-art.
Impressively, our SR with little computing overhead boosted the large-scale network OCRNet by a clear margin of $1.4\%$ mIoU on the \emph{val} set and $0.9\%$ mIoU on the \emph{test} set of CS, and $0.8\%$ on the more challenging PAC.
%%

%-----------------------------------------------------------------
\begin{table}[t]
\small
\begin{center}
\renewcommand\arraystretch{1.1}
\setlength{\tabcolsep}{5pt}{
\begin{tabular}{r|c|cc}
\multicolumn{1}{r|}{Methods}& Backbone & \emph{val} (\%) & \emph{test} (\%)\\
\hline \hline
% SVCNet~\cite{ding2019semantic} & ResNet-101 & -- & 81.0  \\
%ANN~\cite{zhu2019asymmetric} & ResNet-101 & -- & 81.3  \\
%CCNet~\cite{huang2019ccnet} & ResNet-101 & 81.3  & 81.4 \\
DANet~\cite{fu2019dual} & ResNet-101 & 81.5  & 81.5 \\
CDGCNet~\cite{hu2020class} & ResNet-101 & 81.9  & -- \\
HRNet~\cite{WangSCJDZLMTWLX19} & HRNetV2-W48 & 81.1  & 81.6  \\
\hline
DBES~\cite{li2020improving} & ResNet-101 & 81.3$^\flat$ & 81.5$^\flat$ \\
DBES+SR & ResNet-101 & 82.0$_{\color{red}{ +0.7}}$ & 82.1$_{\color{red}{ +0.6}}$ \\
% OCRNet~\cite{YuanCW19} & HRNetV2-W48 & 80.7$^\flat$ & 81.8$^\flat$ \\
% OCRNet+SR & HRNetV2-W48 & 82.1$_{\color{red}{ +1.4}}$ & 82.7$_{\color{red}{ +0.9}}$ \\
% \cdashline{1-4}[0.8pt/2pt]
OCRNet~\cite{YuanCW19} & HRNetV2-W48 & 80.7$^\flat$ & 81.8$^\flat$ \\
OCRNet+SR & HRNetV2-W48 & 82.1$_{\color{red}{ +1.4}}$ & 82.7$_{\color{red}{ +0.9}}$
% DBES~\cite{li2020improving} & ResNet-101 & 81.3$^\flat$ & 81.5$^\flat$ \\
% DBES+SR & ResNet-101 & 82.0$_{\color{red}{ +0.7}}$ & 82.1$_{\color{red}{ +0.6}}$
\end{tabular}}
\vspace{1mm}
\caption{Result comparison (mIoU) with the state-of-the-arts on Cityscapes~\cite{cordts2016cityscapes} using pixel-level labels. ``--'' denotes there are no reported results in the original papers. $\flat$ means that this is our re-implemented result. ``+SR'' means applying SR loss (Eq.~\ref{eq5}) function to train the models.}
\label{tab4}
\end{center}
\vspace{-5mm}
\end{table}
%-----------------------------------------------------------------
\begin{table}[t]
\small
\begin{center}
\renewcommand\arraystretch{1.1}
\setlength{\tabcolsep}{6pt}{
\begin{tabular}{r|c|c}
\multicolumn{1}{r|}{Methods}& Backbone & \emph{test} (\%)\\
\hline \hline
% BFP~\cite{ding2019boundary} & ResNet-101 & 53.6  \\
% DGCNet~\cite{zhang2019dual} & ResNet-101 & 53.7  \\
% HRNet~\cite{WangSCJDZLMTWLX19} & HRNetV2-W48 & 54.0  \\
CFNet~\cite{zhang2019co} & ResNet-101 & 54.0  \\
ACNet~\cite{fu2019adaptive} & ResNet-101 & 54.1  \\
APCNet~\cite{he2019adaptive} & ResNet-101 & 54.7 \\
\hline
DBES~\cite{li2020improving} & ResNet-101 & 54.3 \\
DBES+SR~\cite{li2020improving} & ResNet-101 & 55.3$_{\color{red}{ +1.0}}$ \\
OCRNet~\cite{YuanCW19} & HRNetV2-W48 & 54.9$^\flat$  \\
OCRNet+SR & HRNetV2-W48 & 55.7$_{\color{red}{ +0.8}}$
\end{tabular}}
\vspace{1mm}
\caption{Result comparison (mIoU, \%) with the state-of-the-arts on PASCAL-Context~\cite{mottaghi2014role} using pixel-level labels. }
\label{tab5}
\end{center}
\vspace{-6mm}
\end{table}
%------------------------------------------------------------------------------------
%------------------------------------------------------------------------------------
\begin{figure*}[t]
\centering
\includegraphics[width=.9\textwidth]{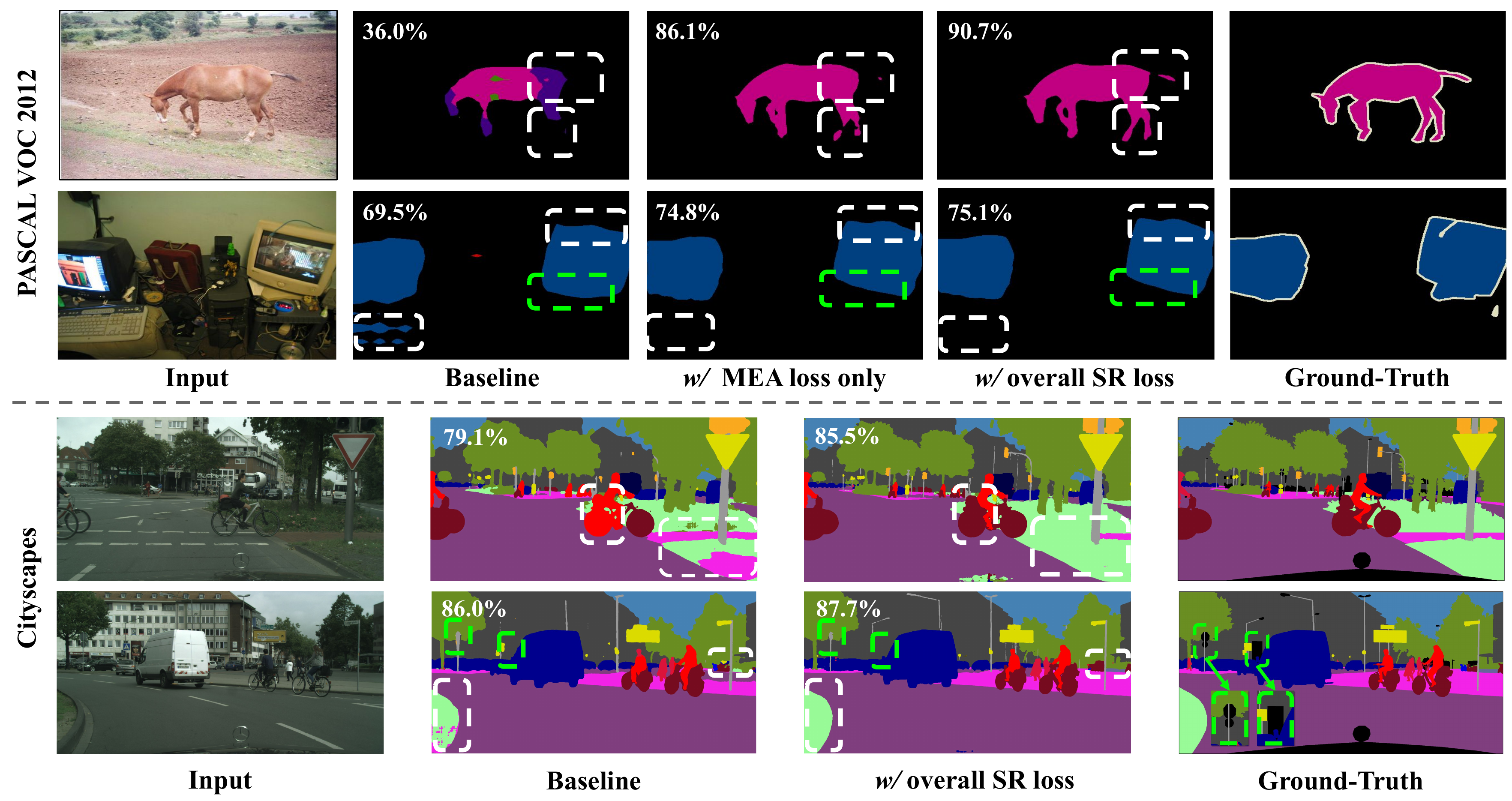}
\caption{Visualization results for WSSS on the \emph{val} set of PASCAL VOC 2012 (PC)~\cite{everingham2010pascal} using CONTA~\cite{zhang2020causal}+SPGNet~\cite{cheng2019spgnet} as Baseline, and FSSS on the \emph{val} set of Cityscapes (CS)~\cite{cordts2016cityscapes} using OCRNet~\cite{YuanCW19} as Baseline.
% ``+'' means adding the loss(es) to the model shown on its left column.
``\emph{{w}}/ overall SR loss'' means applying MEA loss, SR-F loss, and SR-L loss on the Baseline.
The mIoU is shown on each result image.
The white dashed boxes highlight the revised regions by our methods.
The green dashed boxes highlight the failure cases of both baseline and our models.
% segmentation regions.
}
\label{fig4}
\vspace{-4mm}
\end{figure*}
%------------------------------------------------------------------------------------

\myparagraph{Visualizations.}
In Figure~\ref{fig4}, we visualize four segmentation samples on PC (val) and CS (val) datasets. The top two show the results of WSSS models, and the bottom two for FSSS models.
%%
% On the upper block, we show the examples of baseline (\ie, CONTA~\cite{zhang2020causal}+SPGNet~\cite{cheng2019spgnet}), baseline model with MAE loss, and baseline model with all SR loss terms.
%%
From PC samples, we can see that many of the failure regions (highlighted with white dash boxes) using baseline models were corrected by adding MEA losses. This is a gain of strengthening the semantics and details on every individual layer of the model.
On the top of it, using the overall SR loss makes the models more effective to mark out minor object parts,
%\eg, the ``legs'' and ``tail'' of the ``horse'' and the ``foot'' of the ``person''.
\eg, ``horse legs'' and ``horse tail''.
% Consistent results can be found in the CS examples.
%% {\color{blue}}stands
On both datasets, we also saw some failure cases.
% {\color{blue}
For example when segmenting ``monitors'',
all models are missing ``neck'' regions (see green dashed boxes).
% the ``stand'' regions are still lost in the segmentation mask.
We think the reason is that the pseudo mask module (which is basically a classification model) in WSSS rarely attends to ``monitor neck'' when training the classifier of ``monitor''.
%
% WSSS methods haven't contain such information in pseudo-masks.
Another failure case is on the second row of CS: tiny objects such as distant traffic signs are missing. We believe those signs are difficult for human eyes, not to mention for machine models trained on $969\times 969$ images.

% The reason may be that the regulation-based method is still insufficient, and some task-specific methods are further needed in segmentation.
%, but intriguingly, a lot of them are due to the issues of manual annotation.
% present two failure examples respectively in the bottom row of two blocks.
%
% Comparing our overall results to ground truth, we can see that the yellow regions are missing
% For example, on the second row, our method segmented out the ``sofa'' region almost perfectly but this could not be counted in the mIoU as the ground truth of corresponding region was given as ``unknown''.
%
% The reason for these prediction failures is that these areas are not accurately annotated in the dataset, \eg, the sofa area and the car logo area are both annotated as the ``unknown'' category.
%------------------------------------------------------------------------------------
% \input{latex/sections/6_experiments_2}
%------------------------------------------------------------------------------------
\section{Conclusion}
We started by seeking reasons for two major failure cases in SS.
We found that it is either the overuse or underuse of the semantics and visual details across different layers. To this end, we proposed three ``shallow to deep and back'' regulation operations: MEA --- regulating the predictions of each pair with ground-truth labels; SR-F --- regulating deeper-layer segmenters by the shallowest layer segmenter; and SR-L --- regulating shallow-layer classifiers by the deepest layer classifier.
Experimental results on both WSSS and FSSS demonstrated that our SR loss can bring continuous performance gains with little computational overhead.
In the future, we will further investigate some new regulation methods on SS, as well as how to apply existing regulations to solve similar problems in other tasks, such as video event classification and action recognition.
%------------------------------------------------------------------------------------
\section*{Acknowledgements}
This work was partly supported by the National Key Research and Development Program of China under Grant 2018AAA0102002, the National Natural Science Foundation of China under Grant 61732007, A*STAR under its AME YIRG Grant (Project No. A20E6c0101), and the Singapore Ministry of Education (MOE) Academic Research Fund (AcRF) Tier 2 Grant.
%------------------------------------------------------------------------------------
{\small
\bibliographystyle{ieee_fullname}
\bibliography{egbib}
}
%------------------------------------------------------------------------------------
\end{document}